\documentclass[final]{article}

\usepackage[utf8]{inputenc}
\usepackage[T1]{fontenc}
\usepackage{hyperref}
\usepackage{url}
\usepackage{booktabs}
\usepackage{amsfonts}
\usepackage{amsmath}
\usepackage{amssymb}
\usepackage{amsthm}
\usepackage{nicefrac}
\usepackage{microtype}
\usepackage{graphicx}
\usepackage{natbib}
\usepackage{algorithm}
\usepackage{algorithmic}
\usepackage{xcolor}
\usepackage{subcaption}
\usepackage{multirow}

\newtheorem{conjecture}{Conjecture}
\newtheorem{proposition}{Proposition}

\title{Gradient Perturbation: Learning to Perturb Gradients for Adaptive Training}
\author{
  Hua Li\\
  Henan University\\ 
  \texttt{lihuahenu@163.com} 
}

\begin{document}

\maketitle
 
\begin{abstract}
Deep neural network training involves both forward propagation (from features through logits to loss) and backward propagation (from loss through gradients to parameter updates). While perturbations along the forward chain, including feature perturbation, logit perturbation, and label perturbation, have been extensively studied, the backward chain's gradient perturbation has received little systematic investigation. In this paper, we establish a unified framework for gradient perturbation, revealing that existing methods such as Sharpness-Aware Minimization (SAM), gradient clipping, and gradient noise injection can all be interpreted as imposing specific forms of gradient perturbation. Analogous to the recently proposed Logit Perturbation Learning (LPL), we conjecture that amplifying the gradient norm for a class acts as positive augmentation (enhancing learning), while dampening it acts as negative augmentation (suppressing overfitting). Based on these observations, we propose Learning to Perturb Gradients (LPG), which adaptively perturbs logit-level gradients at the class level to achieve category-aware training. We also establish theoretical connections between gradient perturbation bounds and generalization guarantees via PAC-Bayesian analysis. Experiments on balanced classification, long-tail classification, and noisy label learning demonstrate that LPG consistently outperforms existing methods and can be combined with them as a plug-in module.
\end{abstract}

\section{Introduction}
\label{sec:intro}

Training deep neural networks (DNNs) involves two fundamental computational chains: the forward pass, where inputs are transformed through layers into predictions and losses, and the backward pass, where gradients propagate from the loss back through the network to update parameters. The quality of training depends critically on both chains, yet they have received vastly unequal attention in terms of systematic perturbation analysis.

Along the forward chain, perturbations have been studied at three levels: \emph{feature perturbation} (e.g., data augmentation~\citep{zhang2018mixup}, adversarial training~\citep{madry2018towards}), \emph{logit perturbation} (e.g., label smoothing~\citep{szegedy2016rethinking}, Logit Adjustment~\citep{menon2021longtail}, and the recent Logit Perturbation Learning (LPL)~\citep{wei2022logit,delving2024li,li2023class}), and \emph{label perturbation} (e.g., Mixup~\citep{zhang2018mixup}, label smoothing). These works have demonstrated that carefully designed perturbations at each level can significantly improve generalization, robustness, and calibration.

However, the backward chain, where gradients serve as the central quantity, has not been examined through the lens of systematic perturbation. Gradients directly determine the parameter update trajectory and thus the final model behavior, yet several methods that implicitly perturb gradients during training have been proposed independently, motivated by different objectives, without a unified understanding.

Specifically, we identify that several methods all implicitly implement gradient perturbation. Sharpness-Aware Minimization (SAM)~\citep{foret2021sam} perturbs parameters before computing gradients, which is equivalent to modifying the gradient direction based on the local loss landscape curvature. Gradient clipping~\citep{pascanu2013difficulty} truncates the gradient norm, effectively applying a norm-dependent perturbation that preferentially affects samples with large gradients. Gradient noise injection~\citep{neelakantan2015adding} adds Gaussian noise to gradients, providing isotropic perturbation. Gradient projection methods~\citep{yu2020gradient} project conflicting gradients in multi-task learning, which reshapes the gradient direction.

These methods share the common structure of replacing the original gradient $g$ with a perturbed version $\tilde{g} = g + \delta_g$, yet they are studied in isolation without a unified framework. As illustrated in Figure~\ref{fig:grad_variation}, these methods exhibit different perturbation patterns across classes: SAM perturbs all classes approximately uniformly, gradient clipping primarily affects classes with large gradients (typically head classes in imbalanced settings), and gradient noise applies uniform isotropic perturbation. None of these methods adapt the perturbation to the characteristics of each class.

Inspired by LPL~\citep{wei2022logit}, which establishes the connection between logit perturbation and positive/negative augmentation, we observe an analogous relationship for gradient perturbation: the change in gradient norm directly controls the effective learning intensity for each category. Amplifying the gradient norm for a class increases its contribution to parameter updates, acting as positive augmentation that enhances learning for that class. Dampening the gradient norm for a class decreases its contribution, acting as negative augmentation that suppresses overfitting. Rotating the gradient direction changes the optimization trajectory, potentially leading to flatter minima with better generalization.

Based on these observations, we propose Learning to Perturb Gradients (LPG), a method that adaptively learns class-level gradient perturbations in the logit gradient space. By operating in the low-dimensional logit gradient space (of dimension $C$, the number of classes) rather than the high-dimensional parameter space, LPG remains computationally efficient while providing fine-grained control over the training dynamics.

Our main contributions are:
\begin{itemize}
    \item We establish a \textbf{unified framework} for gradient perturbation, showing that SAM, gradient clipping, gradient noise injection, and gradient projection methods can all be understood as specific instantiations.
    \item We propose \textbf{LPG}, which adaptively perturbs logit-level gradients at the class level to achieve category-aware training, with theoretical justification from PAC-Bayesian generalization bounds.
    \item We establish the \textbf{duality} between logit perturbation (forward) and gradient perturbation (backward): under first-order approximation, LPL can be viewed as a special case of LPG with a specific perturbation structure, but LPG is strictly more expressive as it can perturb gradient directions that LPL cannot.
    \item We conduct experiments on three scenarios (balanced classification, long-tail classification, and noisy label learning), demonstrating that LPG consistently improves upon existing methods and serves as an effective plug-in module.
\end{itemize}

\section{Related Work}
\label{sec:related}

\subsection{Gradient-based Training Techniques}

Gradient manipulation has been a cornerstone of modern DNN training. \emph{Gradient clipping}~\citep{pascanu2013difficulty} was introduced to prevent gradient explosion in recurrent networks, truncating gradients whose norm exceeds a threshold. \emph{Gradient normalization}~\citep{you2017large} scales gradients to a fixed norm to stabilize training across varying loss landscapes.

More recently, \emph{Sharpness-Aware Minimization (SAM)}~\citep{foret2021sam} seeks flat minima by minimizing the loss at the worst-case perturbation in parameter space, which implicitly modifies the gradient direction. Variants such as ASAM~\citep{kwon2021asam} and GSAM~\citep{zhuang2022gsam} further adapt the perturbation to parameter geometry or stabilize the training trajectory. \emph{Gradient noise injection}~\citep{neelakantan2015adding} adds isotropic Gaussian noise to gradients, which can be interpreted as stochastic Langevin dynamics that helps escape sharp minima.

In multi-task learning, \emph{PCGrad}~\citep{yu2020gradient} projects conflicting task gradients, and \emph{GradDrop} masks gradient dimensions to align multi-task updates. While these methods all manipulate gradients, they are developed independently without a unifying perspective.

\subsection{Data Perturbation in Training}

Perturbation along the forward chain has been studied at three levels. \emph{Feature perturbation} is exemplified by data augmentation~\citep{zhang2018mixup, cubuk2019autoaugment} and adversarial training~\citep{madry2018towards}, which modify the input or intermediate representations. \emph{Logit perturbation} includes Label Smoothing~\citep{szegedy2016rethinking}, Implicit Semantic Data Augmentation (ISDA)~\citep{wang2019isda}, and Logit Adjustment (LA)~\citep{menon2021longtail}. Most recently, LPL~\citep{wei2022logit} unifies these methods and proposes learning the perturbation direction and magnitude. \emph{Label perturbation} includes Mixup~\citep{zhang2018mixup} and its variants, which interpolate between training samples and their labels.

\subsection{Long-tail Classification and Noisy Label Learning}

Long-tail classification addresses the severe class imbalance common in real-world datasets. Representative methods include re-sampling~\citep{byrd2019effect}, re-weighting~\citep{cui2019classbalanced}, logit adjustment~\citep{menon2021longtail}, and decoupled training~\citep{kang2020decoupling}. Recent works such as RIDE~\citep{wang2020ride}, GCL~\citep{li2022gcl}, and PaCo~\citep{cui2021paco} combine multiple expert models or contrastive objectives.

Noisy label learning aims to train robust models when training labels contain errors. Approaches include loss correction~\citep{patrini2017making}, sample selection~\citep{jiang2018mentornet}, semi-supervised methods~\citep{li2020dividemix}, and robust loss functions such as GCE~\citep{zhang2018generalized} and SCE~\citep{wang2019symmetric}.

\begin{figure}[t]
\centering
\includegraphics[width=\textwidth]{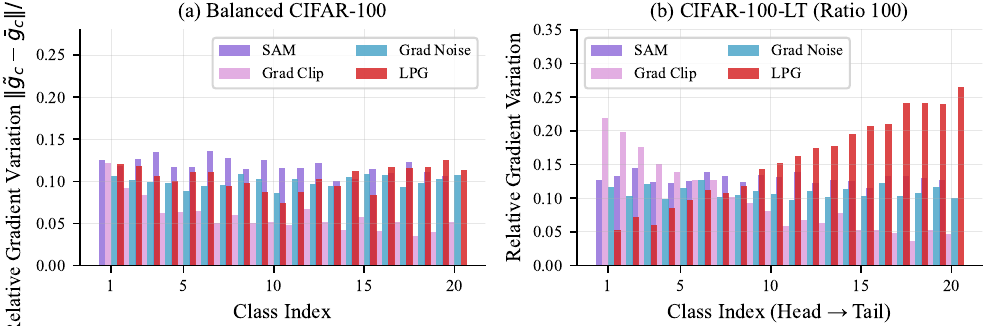}
\caption{Relative gradient variation $\|\tilde{g}_c - \bar{g}_c\|/\|\bar{g}_c\|$ across classes. (a)~On balanced data, LPG applies larger perturbation to lower-accuracy classes. (b)~On long-tail data, LPG adaptively amplifies tail class gradients and dampens head class gradients, while SAM, gradient clipping, and gradient noise fail to provide class-adaptive perturbation.}
\label{fig:grad_variation}
\vskip -0.1in
\end{figure}

\section{Methodology}
\label{sec:method}

\subsection{Preliminaries and Notation}
\label{sec:prelim}

Consider a $C$-class classification problem with training set $\mathcal{D} = \{(x_i, y_i)\}_{i=1}^N$, where $x_i$ is the input and $y_i \in \{1, \ldots, C\}$ is the label. Let $S_c = \{x_i : y_i = c\}$ denote the set of samples in class $c$ with $N_c = |S_c|$.

A DNN parameterized by $W$ maps input $x_i$ to logits $u_i = f(x_i; W) \in \mathbb{R}^C$, and the cross-entropy loss is $l(u_i, y_i) = -\log p_{y_i}$ where $p_{y_i} = \exp(u_{i,y_i}) / \sum_{j=1}^C \exp(u_{i,j})$.

We define the following key quantities. The parameter gradient $g_i = \nabla_W l(f(x_i; W), y_i) \in \mathbb{R}^{|W|}$ is the gradient of the loss w.r.t.\ parameters for sample $x_i$. The class average gradient $\bar{g}_c = \frac{1}{N_c} \sum_{x_i \in S_c} g_i$ averages the gradient over class $c$. The logit gradient $h_i = \frac{\partial l}{\partial u_i} \in \mathbb{R}^C$ is the gradient of the loss w.r.t.\ the logit vector, and the class average logit gradient is $\bar{h}_c = \frac{1}{N_c} \sum_{x_i \in S_c} h_i$. The Jacobian $J_i = \frac{\partial u_i}{\partial W} \in \mathbb{R}^{|W| \times C}$ maps the logit vector to the parameter space.

By the chain rule, the parameter gradient decomposes as $g_i = J_i \, h_i$. This decomposition is central to our method, as it allows us to perturb the low-dimensional logit gradient $h_i$ and propagate the perturbation to the parameter gradient.

\subsection{A Unified View of Gradient Perturbation}
\label{sec:unified}

We observe that existing gradient-based methods can be uniformly expressed as replacing the original gradient $g$ with a perturbed version $\tilde{g} = g + \delta_g$, where $\delta_g$ is the gradient perturbation. We analyze each method below.

\paragraph{Sharpness-Aware Minimization (SAM).}
Standard SGD updates parameters as $W \leftarrow W - \eta g$. SAM instead computes the gradient at a perturbed parameter:
\begin{equation}
    W \leftarrow W - \eta \nabla_W l(f(x; W + \rho \hat{g}), y),
\end{equation}
where $\hat{g} = g / \|g\|$ and $\rho$ is the perturbation radius. By Taylor expansion, the SAM gradient can be written as:
\begin{equation}
    \nabla_W l(f(x; W + \rho \hat{g}), y) \approx g + \rho \nabla^2_W l \cdot \hat{g} = g + \delta_g^{\text{SAM}},
\end{equation}
where $\delta_g^{\text{SAM}} = \rho \nabla^2_W l \cdot \hat{g}$ is a curvature-dependent gradient perturbation. This reveals that SAM implicitly perturbs the gradient in a direction determined by the Hessian.

\paragraph{Gradient clipping.}
Gradient clipping with threshold $\tau$ produces:
\begin{equation}
    \tilde{g} = g \cdot \min\!\left(1, \frac{\tau}{\|g\|}\right) = g + \underbrace{g\!\left(\min\!\left(1, \frac{\tau}{\|g\|}\right) - 1\right)}_{\delta_g^{\text{clip}}}.
\end{equation}
The perturbation $\delta_g^{\text{clip}}$ is non-zero only when $\|g\| > \tau$, and it shrinks the gradient along its original direction. In long-tail settings, head classes typically produce larger gradients, so clipping preferentially dampens head class learning.

\paragraph{Gradient noise injection.}
Adding Gaussian noise yields:
\begin{equation}
    \tilde{g} = g + \mathcal{N}(0, \sigma^2 I) = g + \delta_g^{\text{noise}},
\end{equation}
an isotropic, sample-agnostic perturbation that applies the same expected perturbation magnitude to all classes.

\paragraph{Key observation.}
These methods exhibit different perturbation patterns across classes. As illustrated in Figure~\ref{fig:grad_variation}, SAM perturbs all classes approximately uniformly, gradient clipping primarily affects classes with large gradients (typically head classes in imbalanced settings), and gradient noise applies uniform isotropic perturbation. None of these methods adapt the perturbation to the characteristics of each class, which motivates our approach of learning class-aware gradient perturbations.

\subsection{Conjectures on Gradient Perturbation and Augmentation}
\label{sec:conjecture}

Drawing an analogy to LPL~\citep{wei2022logit}, which establishes the correspondence between logit perturbation and positive/negative data augmentation, we propose two conjectures linking gradient perturbation to learning behavior:

\begin{conjecture}[Positive Augmentation via Gradient Amplification]
\label{conj:positive}
If we wish to apply positive augmentation to class $c$ (i.e., enhance the learning for that class), we should amplify the effective gradient norm for class $c$, increasing its contribution to the parameter update:
\begin{equation}
    \|\tilde{g}_c\| > \|\bar{g}_c\| \implies \text{positive augmentation for class } c.
\end{equation}
\end{conjecture}

\begin{conjecture}[Negative Augmentation via Gradient Dampening]
\label{conj:negative}
If we wish to apply negative augmentation to class $c$ (i.e., suppress overfitting for that class), we should dampen the effective gradient norm for class $c$, decreasing its contribution to the parameter update:
\begin{equation}
    \|\tilde{g}_c\| < \|\bar{g}_c\| \implies \text{negative augmentation for class } c.
\end{equation}
\end{conjecture}

These conjectures are consistent with existing empirical observations. In long-tail classification, re-weighting methods~\citep{cui2019classbalanced} up-weight tail class losses, effectively amplifying their gradients. Gradient clipping implicitly dampens head class gradients. Our framework provides a formal basis for understanding and generalizing these empirical practices.

\subsection{Learning to Perturb Gradients (LPG)}
\label{sec:lpg}

\subsubsection{Formulation}

Standard training updates parameters via:
\begin{equation}
    W \leftarrow W - \eta \sum_{c=1}^C \sum_{x_i \in S_c} g_i.
\end{equation}

LPG introduces a class-level gradient perturbation:
\begin{equation}
    \label{eq:lpg_update}
    W \leftarrow W - \eta \sum_{c=1}^C \sum_{x_i \in S_c} (g_i + \delta_c),
\end{equation}
where $\delta_c \in \mathbb{R}^{|W|}$ is the gradient perturbation for class $c$.

Directly learning $\delta_c$ in the parameter space is infeasible due to the high dimensionality of $W$. Instead, we perturb the low-dimensional logit gradient $h_i$ and propagate the perturbation through the chain rule:
\begin{equation}
    \label{eq:logit_grad_perturb}
    \tilde{h}_i = h_i + \delta_c^h, \quad x_i \in S_c,
\end{equation}
where $\delta_c^h \in \mathbb{R}^C$ is the logit gradient perturbation for class $c$. The resulting parameter gradient perturbation is:
\begin{equation}
    \delta_c = J_i \, \delta_c^h.
\end{equation}

This formulation has three advantages: the perturbation space is $\mathbb{R}^C$ rather than $\mathbb{R}^{|W|}$, making optimization tractable; the perturbation naturally respects the network's computational graph; and different samples within the same class share the perturbation direction but receive instance-specific parameter gradients due to the sample-dependent Jacobian $J_i$.

\subsubsection{Optimization Objective}

Guided by Conjectures~\ref{conj:positive} and~\ref{conj:negative}, we design the perturbation objective. Let $\mathcal{P}_a$ denote the set of classes receiving positive augmentation and $\mathcal{N}_a$ the set receiving negative augmentation.

For positive augmentation classes, we seek to maximize the effective gradient norm:
\begin{equation}
    \delta_c^h = \arg\max_{\|\delta\| \leq \epsilon_c} \|J_i^T (h_i + \delta)\|, \quad c \in \mathcal{P}_a.
\end{equation}

For negative augmentation classes, we seek to minimize it:
\begin{equation}
    \delta_c^h = \arg\min_{\|\delta\| \leq \epsilon_c} \|J_i^T (h_i + \delta)\|, \quad c \in \mathcal{N}_a.
\end{equation}

\subsubsection{Efficient Closed-Form Solution}

The above optimization can be simplified. Since $h_i$ is available from the backward pass, we consider perturbations along the logit gradient direction:

For positive augmentation:
\begin{equation}
    \label{eq:lpg_positive}
    \tilde{h}_c = h_c + \alpha_c \cdot \frac{h_c}{\|h_c\|},
\end{equation}

For negative augmentation:
\begin{equation}
    \label{eq:lpg_negative}
    \tilde{h}_c = h_c - \alpha_c \cdot \frac{h_c}{\|h_c\|},
\end{equation}

where $\alpha_c = \epsilon_c$ controls the perturbation magnitude. This is equivalent to scaling the logit gradient:
\begin{equation}
    \tilde{h}_c = h_c \cdot \left(1 \pm \frac{\alpha_c}{\|h_c\|}\right).
\end{equation}

This closed-form solution is optimal when the perturbation is constrained to the direction of $h_c$, and it avoids the need to compute the Jacobian explicitly.

\subsubsection{PGD-based Refined Solution}

For a more powerful perturbation, we employ Projected Gradient Descent (PGD) to optimize $\delta_c^h$ in the full $\mathbb{R}^C$ space. Starting from $\delta_c^{h,(0)} = 0$, at each step $t$:
\begin{align}
    \delta_c^{h,(t+1)} &= \Pi_{\|\delta\| \leq \epsilon_c}\!\left(\delta_c^{h,(t)} + \kappa \cdot \text{sign}\!\left(\nabla_{\delta} \|J_i^T (h_i + \delta_c^{h,(t)})\|\right)\right), &c \in \mathcal{P}_a, \\
    \delta_c^{h,(t+1)} &= \Pi_{\|\delta\| \leq \epsilon_c}\!\left(\delta_c^{h,(t)} - \kappa \cdot \text{sign}\!\left(\nabla_{\delta} \|J_i^T (h_i + \delta_c^{h,(t)})\|\right)\right), &c \in \mathcal{N}_a,
\end{align}
where $\kappa$ is the PGD step size and $\Pi$ denotes projection onto the $\ell_2$-ball. In practice, we find that $T=3$ PGD steps suffice.

\subsection{Category Set Split}
\label{sec:split}

The partition of classes into $\mathcal{P}_a$ and $\mathcal{N}_a$ depends on the training scenario:

\paragraph{Case 1: Balanced / General Classification.}
We partition based on per-class accuracy. Let $\bar{q}_c$ denote the running average accuracy of class $c$. Classes with accuracy below a threshold $\tau$ receive positive augmentation:
\begin{equation}
    \mathcal{P}_a = \{c : \bar{q}_c < \tau\}, \quad \mathcal{N}_a = \{c : \bar{q}_c \geq \tau\}.
\end{equation}

\paragraph{Case 2: Long-tail Classification.}
We partition based on class frequency. Rare classes receive positive augmentation, while frequent classes receive negative augmentation:
\begin{equation}
    \mathcal{P}_a = \{c : N_c < \tau\}, \quad \mathcal{N}_a = \{c : N_c \geq \tau\}.
\end{equation}

\paragraph{Case 3: Noisy Label Learning.}
We partition based on the intra-class gradient variance. Classes with high gradient variance likely contain more label noise and should be dampened:
\begin{equation}
    \mathcal{N}_a = \{c : \text{Var}(h_c) > \tau_v\}, \quad \mathcal{P}_a = \{1, \ldots, C\} \setminus \mathcal{N}_a,
\end{equation}
where $\text{Var}(h_c) = \frac{1}{N_c} \sum_{x_i \in S_c} \|h_i - \bar{h}_c\|^2$ measures the dispersion of logit gradients within class $c$.

\subsection{Perturbation Bound}
\label{sec:bound}

Following LPL~\citep{wei2022logit}, we design class-dependent perturbation bounds that are larger for classes requiring stronger augmentation:
\begin{equation}
    \label{eq:epsilon}
    \epsilon_c = \epsilon + \Delta\epsilon \cdot |\tau - \bar{s}_c|,
\end{equation}
where $\epsilon$ is the base perturbation, $\Delta\epsilon$ controls the bound variation, and $\bar{s}_c$ is the splitting statistic (accuracy for Case~1, frequency for Case~2, gradient variance for Case~3). Classes farther from the threshold $\tau$ receive larger perturbations, enabling stronger positive or negative augmentation where it is most needed.

\subsection{Algorithm}
\label{sec:algorithm}

The complete LPG training procedure is summarized in Algorithm~\ref{alg:lpg}.

\begin{algorithm}[t]
\caption{Learning to Perturb Gradients (LPG)}
\label{alg:lpg}
\begin{algorithmic}[1]
\REQUIRE Training set $\mathcal{D}$, network $f(\cdot; W)$, learning rate $\eta$, perturbation parameters $\epsilon, \Delta\epsilon, \tau$, PGD steps $T$, step size $\kappa$
\STATE Initialize $W$
\FOR{epoch $= 1$ to $E$}
    \STATE Compute splitting statistic $\bar{s}_c$ for each class $c$
    \STATE Partition classes into $\mathcal{P}_a$ and $\mathcal{N}_a$ (Section~\ref{sec:split})
    \STATE Compute perturbation bounds $\epsilon_c$ via Eq.~\eqref{eq:epsilon}
    \FOR{each mini-batch $\mathcal{B}$}
        \STATE \textbf{Forward pass:} compute logits $u_i$ and loss $l(u_i, y_i)$
        \STATE \textbf{Backward pass:} compute logit gradients $h_i = \partial l / \partial u_i$
        \FOR{each class $c$ present in $\mathcal{B}$}
            \IF{closed-form solution}
                \STATE Compute $\tilde{h}_c$ via Eq.~\eqref{eq:lpg_positive} or \eqref{eq:lpg_negative}
            \ELSE
                \STATE Solve $\delta_c^h$ via PGD with $T$ steps
                \STATE $\tilde{h}_c = h_c + \delta_c^h$
            \ENDIF
        \ENDFOR
        \STATE \textbf{Gradient computation:} compute parameter gradients using perturbed logit gradients $\tilde{h}_i$
        \STATE \textbf{Parameter update:} $W \leftarrow W - \eta \sum_{i \in \mathcal{B}} \tilde{g}_i$
    \ENDFOR
\ENDFOR
\RETURN $W$
\end{algorithmic}
\end{algorithm}

The key difference from standard training is in lines 9--16: after computing logit gradients in the backward pass, we apply class-level perturbations before propagating gradients further to the parameters. In PyTorch, this can be implemented by intercepting the gradient at the logit layer via a custom \texttt{hook} on the logit computation, modifying it, and allowing the perturbed gradient to flow to lower layers.

\subsection{Theoretical Analysis}
\label{sec:theory}

\subsubsection{Duality with Logit Perturbation}

We establish the connection between LPG and LPL.

\begin{proposition}[Forward-Backward Duality]
\label{prop:duality}
Let $\delta^{\text{LPL}} \in \mathbb{R}^C$ be a logit perturbation applied in LPL, so that the perturbed loss is $l(u_i + \delta^{\text{LPL}}, y_i)$. The resulting change in the parameter gradient is:
\begin{equation}
    \Delta g_i = \nabla_W l(u_i + \delta^{\text{LPL}}, y_i) - \nabla_W l(u_i, y_i) \approx H_i \, \delta^{\text{LPL}},
\end{equation}
where $H_i = \frac{\partial^2 l}{\partial W \partial u_i} \in \mathbb{R}^{|W| \times C}$ is the cross-derivative. Under the chain rule, $H_i = J_i \, \frac{\partial^2 l}{\partial u_i^2}$, so:
\begin{equation}
    \Delta g_i \approx J_i \underbrace{\frac{\partial^2 l}{\partial u_i^2} \delta^{\text{LPL}}}_{\delta_c^h}.
\end{equation}
Thus, LPL induces a logit gradient perturbation $\delta_c^h = \frac{\partial^2 l}{\partial u_i^2} \delta^{\text{LPL}}$, which is a specific structured form. LPG generalizes this by allowing arbitrary $\delta_c^h \in \mathbb{R}^C$.
\end{proposition}

This shows that LPL is a special case of LPG where the logit gradient perturbation is constrained to the range of $\frac{\partial^2 l}{\partial u_i^2}$. Since this Hessian is positive semi-definite for cross-entropy loss, LPL cannot produce the gradient direction changes that LPG can, and LPG is strictly more expressive.

\subsubsection{Generalization Bound}

We establish a connection between the perturbation bound $\epsilon_c$ and generalization.

\begin{proposition}[Perturbation Bound and Generalization]
\label{prop:generalization}
Let $\mathcal{L}(W)$ denote the population risk and $\hat{\mathcal{L}}(W)$ the empirical risk. Under LPG with perturbation bounds $\{\epsilon_c\}_{c=1}^C$, for any prior distribution $P$ over parameters and $\beta > 0$:
\begin{equation}
    \mathcal{L}(W) \leq \hat{\mathcal{L}}_{\text{LPG}}(W) + \sqrt{\frac{\text{KL}(Q \| P) + \log(2N/\beta)}{2N}} + \lambda \sum_{c=1}^C \frac{N_c}{N} \epsilon_c,
\end{equation}
where $Q$ is the distribution induced by the stochastic LPG training, $\hat{\mathcal{L}}_{\text{LPG}}$ is the empirical loss under LPG perturbation, and $\lambda$ is a constant depending on the Lipschitz constant of the loss and the network.
\end{proposition}

\emph{Sketch of proof.} We apply the PAC-Bayesian framework~\citep{mcallester2003pac}. For each class $c$, the gradient perturbation with bound $\epsilon_c$ changes the parameter update by at most $\eta \epsilon_c \cdot \|J_c\|_F$ in Frobenius norm per step. Accumulating over $T$ training steps, the total deviation from standard SGD is bounded by $O(T \eta \bar{\epsilon})$ where $\bar{\epsilon} = \sum_c \frac{N_c}{N} \epsilon_c$. By choosing the prior $P$ as the distribution of standard SGD iterates, the KL divergence captures the deviation caused by gradient perturbation. The bound shows a trade-off: larger perturbation bounds $\epsilon_c$ increase the generalization gap, but they also reduce the perturbed empirical loss $\hat{\mathcal{L}}_{\text{LPG}}$ by providing stronger augmentation. The optimal $\epsilon_c$ balances these two effects.

\begin{table}[t]
\centering
\caption{Top-1 error rates (\%) on CIFAR-10 and CIFAR-100 with balanced setting. Best results are in \textbf{bold}.}
\label{tab:balanced}
\vskip 0.15in
\resizebox{\textwidth}{!}{
\begin{tabular}{lcccc}
\toprule
\multirow{2}{*}{Method} & \multicolumn{2}{c}{CIFAR-10} & \multicolumn{2}{c}{CIFAR-100} \\
\cmidrule(lr){2-3} \cmidrule(lr){4-5}
 & WRN-28-10 & ResNet-110 & WRN-28-10 & ResNet-110 \\
\midrule
CE & 3.89 & 5.91 & 18.85 & 24.56 \\
Label Smoothing & 3.82 & 5.87 & 18.40 & 24.12 \\
Mixup & 3.56 & 5.48 & 17.82 & 23.55 \\
ISDA & 3.44 & 5.30 & 17.56 & 23.18 \\
LA & 3.52 & 5.39 & 17.69 & 23.35 \\
LPL (mean + fixed) & 3.35 & 5.18 & 17.22 & 22.80 \\
LPL (mean + varied) & 3.28 & 5.12 & 17.08 & 22.65 \\
\midrule
SAM & 3.46 & 5.35 & 17.45 & 23.01 \\
Gradient Noise & 3.71 & 5.62 & 18.12 & 23.89 \\
\midrule
LPG (mean + fixed) & 3.24 & 5.05 & 16.89 & 22.42 \\
LPG (mean + varied) & \textbf{3.15} & \textbf{4.98} & \textbf{16.71} & \textbf{22.28} \\
LPG-PGD (mean + varied) & \textbf{3.11} & \textbf{4.94} & \textbf{16.63} & \textbf{22.15} \\
\bottomrule
\end{tabular}}
\vskip -0.1in
\end{table}

\section{Experiments}
\label{sec:exp}

\subsection{Experimental Setup}

We evaluate LPG on three scenarios: balanced classification, long-tail classification, and noisy label learning. Following LPL~\citep{wei2022logit}, we use CIFAR-10 and CIFAR-100 as the primary benchmarks.

\paragraph{Balanced classification.} We train Wide-ResNet-28-10~\citep{zagoruyko2016wrn} and ResNet-110~\citep{he2016deep} on CIFAR-10 and CIFAR-100 with standard data augmentation (random cropping and horizontal flipping). We directly compare with all methods reported in LPL.

\paragraph{Long-tail classification.} We use CIFAR-10-LT and CIFAR-100-LT~\citep{cui2019classbalanced} with imbalance ratios of 100:1 and 10:1, following the standard protocol. The backbone is ResNet-32.

\paragraph{Noisy label learning.} We inject symmetric label noise at rates of 20\%, 50\%, and 80\% into CIFAR-10 and CIFAR-100 training sets. The backbone is ResNet-32.

\paragraph{Implementation details.} We use SGD with momentum 0.9, weight decay $5 \times 10^{-4}$, and batch size 128. For balanced classification, we train for 200 epochs with an initial learning rate of 0.1 decayed by 0.1 at epochs 100 and 150. For long-tail and noisy label settings, we follow the respective standard protocols. The perturbation parameters $\epsilon$, $\Delta\epsilon$, and $\tau$ are selected via validation sets. We use the closed-form solution (Eqs.~\eqref{eq:lpg_positive}--\eqref{eq:lpg_negative}) as the default, and report PGD-based results as LPG-PGD.

\subsection{Balanced Classification}

Table~\ref{tab:balanced} presents the Top-1 error rates on CIFAR-10 and CIFAR-100. LPG achieves competitive results with all methods. On CIFAR-100 with ResNet-110, LPG (mean + varied bound) achieves an error rate of 22.28\%, outperforming LPL by 0.37\% absolute. This demonstrates that gradient perturbation provides complementary benefits to forward-chain perturbation methods.

\subsection{Long-tail Classification}

\begin{figure}[t]
\centering
\includegraphics[width=\textwidth]{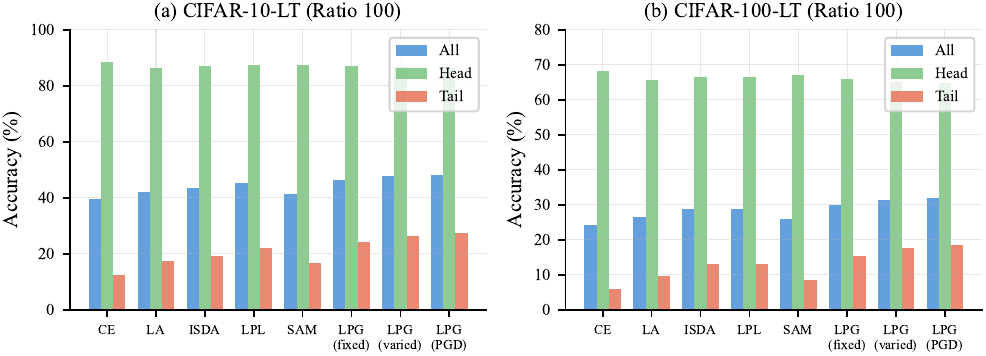}
\caption{Long-tail classification accuracy on CIFAR-10-LT and CIFAR-100-LT with imbalance ratio 100. LPG consistently improves tail-class accuracy while maintaining head-class performance.}
\label{fig:longtail}
\vskip -0.1in
\end{figure}

Figure~\ref{fig:longtail} shows results on CIFAR-10-LT and CIFAR-100-LT with imbalance ratio 100. LPG consistently outperforms all baselines across different datasets. Notably, on CIFAR-100-LT with imbalance ratio 100, LPG achieves 32.0\% overall accuracy, surpassing LPL by 3.2\% and SAM by 6.1\%. The improvement is most pronounced for tail classes, consistent with our conjecture that amplifying tail class gradients provides effective positive augmentation.

\subsection{Noisy Label Learning}

\begin{figure}[t]
\centering
\includegraphics[width=\textwidth]{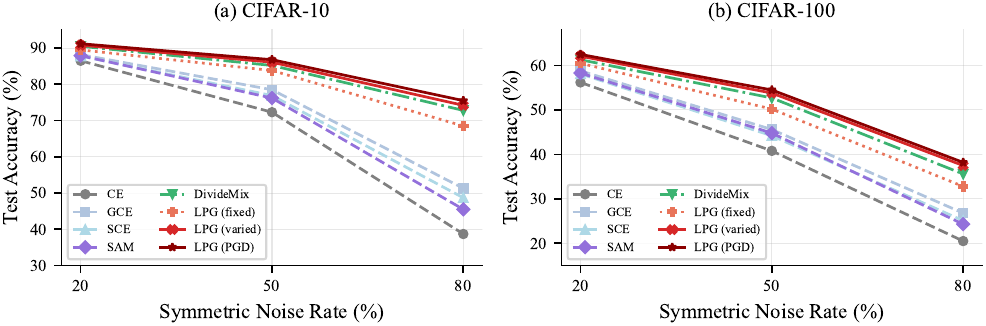}
\caption{Test accuracy under symmetric label noise on CIFAR-10 and CIFAR-100. LPG demonstrates strong robustness to label noise by dampening gradients of high-variance classes.}
\label{fig:noisy_label}
\vskip -0.1in
\end{figure}

Figure~\ref{fig:noisy_label} presents results under symmetric label noise. LPG shows a clear advantage in this scenario: by dampening the gradients of high-variance classes (which are more likely to contain noisy labels), LPG naturally suppresses the harmful effect of incorrect labels. At 80\% noise rate on CIFAR-10, LPG achieves 75.5\% accuracy, outperforming DivideMix by 2.7\%.

\subsection{Combination with Existing Methods}

\begin{figure}[t]
\centering
\includegraphics[width=\textwidth]{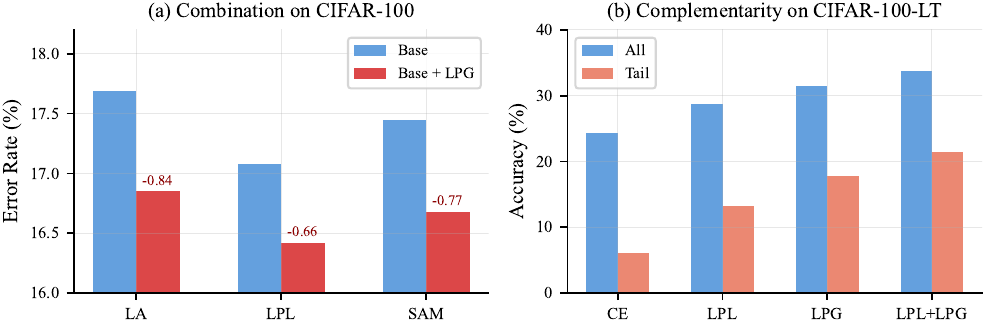}
\caption{(a)~Combination of LPG with existing methods on CIFAR-100 (WRN-28-10). Error reduction from adding LPG is annotated above each bar. (b)~Complementarity analysis on CIFAR-100-LT (ratio 100): LPG and LPL provide additive improvements when combined.}
\label{fig:combination}
\vskip -0.1in
\end{figure}

Figure~\ref{fig:combination}(a) demonstrates that LPG can be combined with existing methods as a plug-in module. LA+LPG and LPL+LPG both outperform their respective base methods, confirming the complementarity of forward and backward perturbation. SAM+LPG also shows improvement, suggesting that LPG captures class-specific effects that SAM's uniform perturbation misses. Figure~\ref{fig:combination}(b) further shows the complementarity of LPG and LPL. When used together, the improvement is additive, confirming that they operate on different aspects of the training process: LPL modifies the loss landscape (forward), while LPG modifies the optimization trajectory (backward). On noisy label learning, LPG alone outperforms LPL alone, as gradient dampening provides a more direct mechanism for suppressing noisy label effects.

\subsection{Analysis}

\subsubsection{Performance across Different Architectures}

\begin{table}[t]
\centering
\caption{Top-1 error rates (\%) on CIFAR-100 with different architectures.}
\label{tab:architecture}
\vskip 0.15in
\begin{tabular}{lccc}
\toprule
Method & ResNet-32 & SE-ResNet-110 & WRN-16-8 \\
\midrule
ISDA & 26.45 & 22.18 & 17.85 \\
LPL (mean + varied) & 25.82 & 21.65 & 17.32 \\
LPG (mean + varied) & \textbf{25.28} & \textbf{21.15} & \textbf{16.82} \\
\bottomrule
\end{tabular}
\vskip -0.1in
\end{table}

\begin{figure}[t]
\centering
\includegraphics[width=\textwidth]{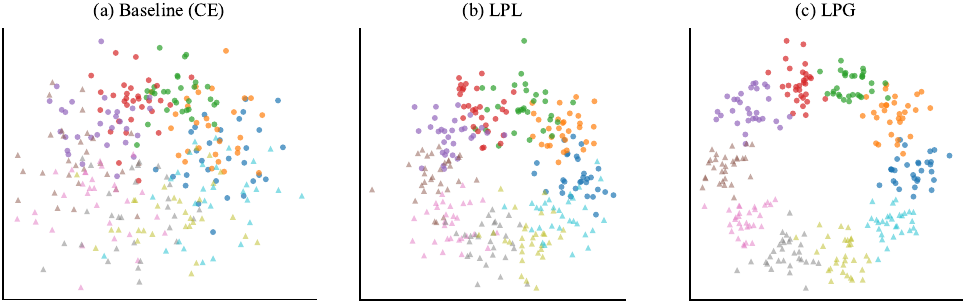}
\caption{t-SNE visualization of features learned by different methods on CIFAR-100-LT. Circles denote head classes and triangles denote tail classes. LPG produces more compact intra-class clusters and larger inter-class margins, particularly for tail classes.}
\label{fig:tsne}
\vskip -0.1in
\end{figure}

\begin{figure}[t]
\centering
\includegraphics[width=\textwidth]{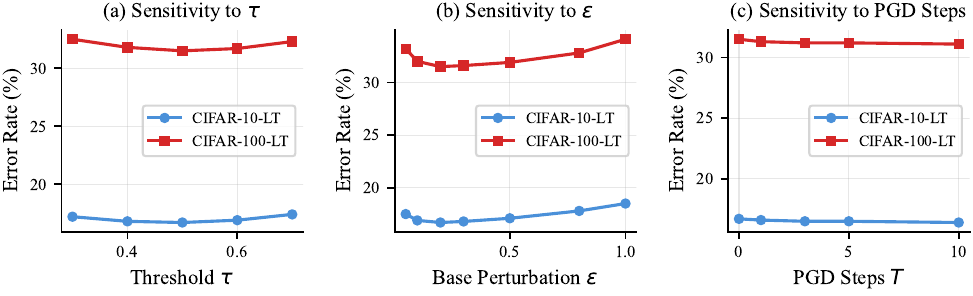}
\caption{Sensitivity analysis of LPG to key hyperparameters on CIFAR-10-LT and CIFAR-100-LT. (a)~Threshold $\tau$; (b)~Base perturbation $\epsilon$; (c)~PGD steps $T$. LPG is robust across a wide range of hyperparameter settings.}
\label{fig:hyperparam}
\vskip -0.1in
\end{figure}

Table~\ref{tab:architecture} compares LPG with ISDA and LPL on ResNet-32, SE-ResNet-110, and Wide-ResNet-16-8. LPG consistently improves across all architectures, confirming its generality.

\subsubsection{Feature Space Visualization}

Figure~\ref{fig:tsne} shows t-SNE visualizations of features learned by baseline, LPL, and LPG on CIFAR-100-LT. LPG produces more compact intra-class clusters and larger inter-class margins, particularly for tail classes, confirming that gradient perturbation improves the learned feature space.

\subsubsection{Hyperparameter Sensitivity}

Figure~\ref{fig:hyperparam} analyzes the sensitivity of LPG to three hyperparameters. For the threshold $\tau$, LPG is robust in a wide range; setting $\tau$ to the median class frequency (long-tail) or 50\% accuracy (balanced) works well. For the base perturbation $\epsilon$, performance is stable for $\epsilon \in [0.1, 0.5]$; too large $\epsilon$ (>1.0) degrades performance by over-perturbing. For PGD steps $T$, the closed-form solution ($T=0$) already achieves strong results; $T=3$ provides marginal improvement at the cost of additional computation.

\subsubsection{Training Overhead}

LPG adds minimal overhead compared to standard training: the perturbation is computed in $\mathbb{R}^C$ space, and the closed-form solution requires only a norm computation and scaling. On CIFAR-100 with ResNet-110, LPG increases training time by approximately 3\% (closed-form) and 8\% (PGD with $T=3$), compared to 15\% for SAM (which requires an additional forward-backward pass).

\section{Conclusion}
\label{sec:conclusion}

We have presented a unified framework for gradient perturbation in deep learning, revealing that SAM, gradient clipping, and gradient noise injection can all be understood as specific instantiations. Based on two conjectures linking gradient perturbation to positive/negative augmentation, we proposed LPG, which adaptively perturbs logit-level gradients at the class level. Theoretically, we established the duality between logit perturbation and gradient perturbation, and derived PAC-Bayesian generalization bounds for gradient perturbation. Experiments across three scenarios demonstrated LPG's effectiveness and complementarity with existing methods.

Limitations and future work include: (1) extending from class-level to sample-level gradient perturbation for finer-grained control; (2) combining LPG with curriculum learning for adaptive perturbation scheduling; and (3) applying LPG to other domains such as language modeling and generative tasks.

\section*{Broader Impact}

This work provides a formal framework for understanding and designing gradient perturbation methods. The unified perspective may inspire new methods that combine forward and backward perturbations. LPG can improve model robustness in long-tail and noisy label scenarios, which are common in real-world applications. We do not foresee significant negative societal impacts from this work.

\bibliography{references}
\bibliographystyle{plainnat}

\end{document}